\PassOptionsToPackage{table}{xcolor}
\documentclass{article}

\usepackage{iclr2027_conference,times}

\usepackage{amsmath,amsfonts,bm}

\def\eqref#1{equation~\ref{#1}}

\def\1{\bm{1}}

\DeclareMathAlphabet{\mathsfit}{\encodingdefault}{\sfdefault}{m}{sl}
\SetMathAlphabet{\mathsfit}{bold}{\encodingdefault}{\sfdefault}{bx}{n}

\usepackage{booktabs}
\usepackage{caption}
\usepackage{flafter}
\usepackage{graphicx}
\usepackage{hyperref}
\usepackage{microtype}
\usepackage{tabularx}
\usepackage{amssymb}
\usepackage{pifont}
\usepackage{placeins}
\usepackage{xcolor}
\usepackage{url}
\usepackage{wrapfig}
\usepackage{enumitem}
\usepackage{fontawesome5}

\definecolor{checkgreen}{HTML}{2F9461}
\definecolor{crossred}{HTML}{C75454}
\definecolor{poseblue}{HTML}{EEF5F9}
\definecolor{tabrule}{HTML}{CBD5DF}
\definecolor{pubgray}{HTML}{747E89}
\definecolor{scoreheat}{HTML}{3182BD}
\newcommand{\cmark}{\textcolor{checkgreen}{\ding{51}}}
\newcommand{\xmark}{\textcolor{crossred}{\ding{55}}}
\newcommand{\posekey}{\begingroup\setlength{\fboxsep}{1pt}\colorbox{poseblue}{Light-blue}\endgroup}
\newcommand{\scorecell}[2]{\cellcolor{scoreheat!#1!white}#2}
\newcommand{\pub}[1]{\hspace{0.25em}{\scriptsize\color{pubgray}[#1]}}

\title{EgoAlign: Situated 3D Understanding without Situation Descriptions}

\author{
Yiyang Li \qquad
Tianyi Ma \qquad
Zehong Wang \qquad
Yanfang Ye$^{*}$ \\
Department of Computer Science and Engineering, University of Notre Dame \\
$^{*}$Corresponding author\\
\texttt{\{yli62,yye7\}@nd.edu}\\
\href{https://github.com/Yiyang-Ian-Li/EgoAlign}{\faGithub\ \texttt{https://github.com/Yiyang-Ian-Li/EgoAlign}}
}

\iclrfinalcopy

\begin{document}

\maketitle
\lhead{Preprint.}

\begin{abstract}
Situated 3D Question Answering requires models to interpret a 3D scene relative to an agent’s pose. Existing methods commonly convey that pose through a natural-language situation description, which may be unavailable or costly to obtain in real-world deployments. While some methods additionally incorporate numerical pose, we find that randomly shuffling the agent pose causes little performance degradation, indicating limited functional dependence on the numerical pose. To address this gap, we introduce EgoAlign, a simple 3D language model that represents a scene in an agent-centric coordinate frame, together with a dedicated egocentric alignment stage comprising four complementary tasks: object category recognition, egocentric direction prediction, metric distance estimation, and spatial relation prediction. This stage teaches the model to consistently interpret scene geometry relative to the agent’s pose before task-specific fine-tuning. EgoAlign achieves the strongest performance on SQA3D, Real-3DQA, and MSQA among compared methods while remaining highly efficient. More importantly, removing textual situation descriptions yields comparable or better downstream performance, whereas shuffling EgoAlign’s numerical pose causes substantial degradation. These results show that EgoAlign establishes agent-centric spatial context directly from numerical pose without requiring textual situation descriptions.
\end{abstract}

\begin{center}
  \includegraphics[width=\textwidth]{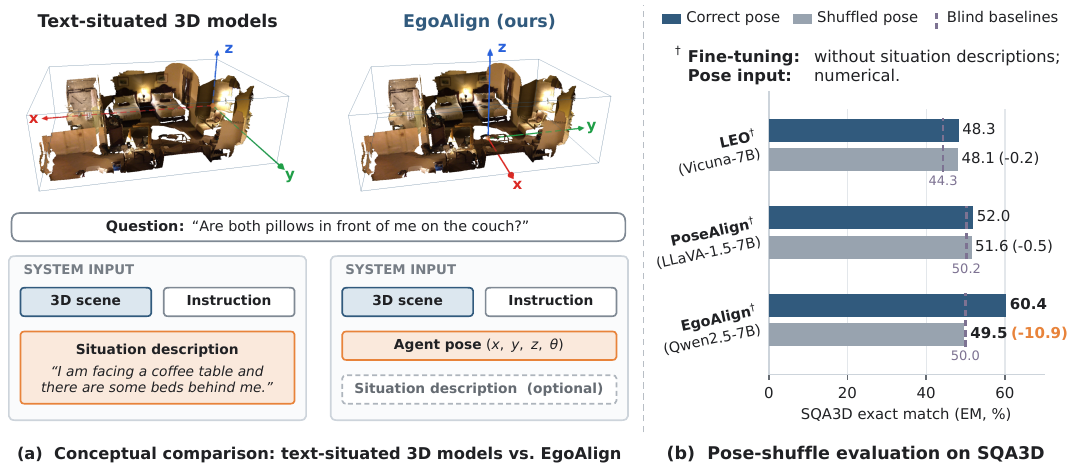}
  \captionsetup{hypcap=false}
  \captionof{figure}{\textbf{Conceptual overview and pose-shuffle evaluation.}
    (a) Text-situated 3D models specify the agent's situation using a natural-language description, whereas EgoAlign constructs its spatial reference frame directly from numerical pose.
    (b) We evaluate pose-conditioned models after fine-tuning on SQA3D without situation descriptions. Blind baseline means model fine-tuned without 3D input. Shuffling poses has little effect on LEO and PoseAlign, but substantially degrades EgoAlign to near the blind-baseline level, showing that EgoAlign’s predictions are strongly dependent on the supplied numerical pose.}
  \label{fig:intro_overview}
\end{center}
\vfill
\clearpage

\section{Introduction}
\label{sec:introduction}

Recent 3D language models have made substantial progress in grounding large language models (LLMs) in 3D environments, enabling tasks such as question answering, grounding, planning, and embodied reasoning~\citep{hong20233d, zhu20233d, chen2024ll3da, huang2023embodied, huang2024chat, deng20253d, zhi2025lscenellm}. For an embodied agent, understanding a scene also requires knowing where the agent is and which direction it faces: whether an object is to the left, in front, or behind depends on the agent’s pose~\citep{zhang2024spartun3d}. Situated 3D benchmarks such as SQA3D and MSQA capture this requirement by conditioning questions on an agent-specific situation~\citep{ma2022sqa3d, linghu2024multi}. Most existing methods represent this situation through natural-language or multimodal descriptions, a formulation that naturally leverages the strong language understanding capabilities of pretrained LLMs~\citep{man2024situational, zhang2024spartun3d, yuan2025empowering}.

However, for a physical agent, numerical pose is a more direct representation of its state.  Localization systems naturally provide position and orientation~\citep{cadena2016past, campos2021orb}, whereas a situation description is an indirect linguistic rendering that must be regenerated as the agent moves~\citep{ma2022sqa3d, liu2026direction}. Ideally, a 3D language model should therefore establish its spatial reference frame directly from numerical pose instead of relying on textual situation descriptions. A few recent models have started to incorporate numerical pose, for example by encoding pose as an additional input or transforming scene geometry into an ego-aligned coordinate frame~\citep{huang2023embodied, linghu2024multi, zhang2024spartun3d, liu2026direction}.

Yet \textbf{incorporating numerical pose is not the same as using it}. QA accuracy alone does not reveal whether a model actually relies on its numerical pose. We test pose dependence directly by shuffling numerical poses while keeping the scene and question unchanged. As shown in Figure~\ref{fig:intro_overview}, even after fine-tuning without situation descriptions, LEO and PoseAlign lose only 0.2 and 0.5 exact-match points, respectively. Their predictions are therefore largely insensitive to whether the supplied pose is correct; more details are provided in Appendix~\ref{app:pose_shuffle_significance}. We refer to this failure mode as \textbf{pose insensitivity}: a model may receive numerical pose, or even transform its input geometry with it, while learning a prediction function that depends little on that pose.

To address this problem, we introduce \textbf{EgoAlign}, a point-cloud language model that makes numerical pose operational at both the representation and supervision levels. At the representation level, it transforms the scene into an agent-centric coordinate frame and explicitly preserves the metric coordinates associated with each point token, so that the LLM receives both semantic and geometric information relative to the current pose. At the supervision level, we introduce an egocentric alignment stage with four complementary tasks: object category recognition, egocentric direction prediction, metric distance estimation, and spatial relation prediction. For each scene configuration, we keep the scene, agent position, and referred objects fixed while varying only the numerical heading across four orientations, creating counterfactual groups that explicitly teach the model how spatial interpretation should change with the agent frame.

EgoAlign achieves the strongest performance on SQA3D, Real-3DQA, and MSQA among compared methods while remaining computationally efficient. More importantly, removing textual situation descriptions yields comparable or better downstream performance, showing that the model can establish its spatial context directly from numerical pose. Under the same pose-shuffle intervention on SQA3D, EgoAlign drops around 10 points to its blind baseline level, while existing pose-conditioned models change little. Its performance also degrades smoothly under increasing yaw perturbations. Together, these results show that EgoAlign is not only accurate but functionally dependent on the numerical pose it receives. Our contributions are: 

\begin{itemize}
    \item We identify \textbf{pose insensitivity} as a hidden failure mode of situated 3D language models.
    \item We propose \textbf{EgoAlign}, an agent-centric point-cloud language model that explicitly preserves metric coordinates at the point-token level.
    \item We introduce an \textbf{egocentric alignment} stage with multi-heading counterfactual supervision that teaches the model to interpret geometry relative to numerical pose.
    \item Across SQA3D, Real-3DQA, and MSQA, EgoAlign demonstrate superior performance without requiring textual situation descriptions and exhibits strong pose dependence.
\end{itemize}

\section{Related Work}
\label{sec:related-work}

\paragraph{Language models for 3D scene understanding.}
Recent 3D language models connect pretrained LLMs with point-cloud or object-level scene representations, enabling tasks such as question answering, captioning, grounding, planning, and embodied interaction~\citep{hong20233d, zhu20233d, wang2023chat, xu2024pointllm}. These methods mainly focus on grounding language in scene content, i.e., recognizing what objects and structures are present and reasoning about their relations~\citep{chen2024grounded, huang2024chat, mei2025perla}. EgoAlign follows this general point-cloud language modeling paradigm, but focuses on a complementary aspect: representing the same 3D scene relative to a changing agent pose. To support such reasoning, we explicitly preserve metric coordinates in the agent-centric frame instead of relying only on compressed semantic features.

\paragraph{Situated 3D understanding.}
Situated 3D understanding extends scene reasoning to a particular agent position and orientation. SQA3D introduces this setting using natural-language situation descriptions, while subsequent works such as SIG3D, SPARTUN3D, and MSQA improve situation grounding, spatial alignment, and multimodal situation modeling~\citep{ma2022sqa3d, man2024situational, zhang2024spartun3d, linghu2024multi}. Most methods still specify the agent state primarily through descriptive language, although a smaller set of models, including LEO, MSR3D, and PoseAlign, also incorporate numerical pose~\citep{huang2023embodied, linghu2024multi, liu2026direction}. EgoAlign differs by treating numerical pose as the primary reference frame and explicitly training the model to make pose-dependent predictions through egocentric alignment supervision.

\section{Method}
\label{sec:method}

\subsection{Preliminary}

Let a 3D scene be a colored point cloud $\mathcal{P}=\{(\mathbf{p}_i,\mathbf{c}_i)\}_{i=1}^{N}$ in the scan's default coordinate system, where $\mathbf{p}_i=(x_i,y_i,z_i)$ and $\mathbf{c}_i=(r_i,g_i,b_i)$ are the XYZ coordinates and RGB color of point $i$, respectively.
For an agent pose with position $\mathbf{t}=(x,y,z)$ and heading $\theta$, we first transform $\mathcal{P}$ into $\mathcal{P}_{\mathrm{ego}}$, where the agent is at the origin with $+x$ forward, $+y$ left, and $+z$ up.
Given an instruction $q$ and an optional situation description $s$ (or $s=\varnothing$ when unavailable), the model predicts
\begin{equation}
  \hat a=f_{\phi}(\mathcal{P}_{\mathrm{ego}},q,s),\quad \mathcal{P}_{\mathrm{ego}}=\{(\widetilde{\mathbf{p}}_i,\mathbf{c}_i)\}_{i=1}^{N},\quad \widetilde{\mathbf{p}}_i=R_z(-\theta)(\mathbf{p}_i-\mathbf{t}).
  \label{eq:agent_transform}
\end{equation}

\subsection{Point-cloud language model}

\begin{wrapfigure}{r}{0.3\textwidth}
  \vspace{-\intextsep}
  \centering
  \includegraphics[width=\linewidth]{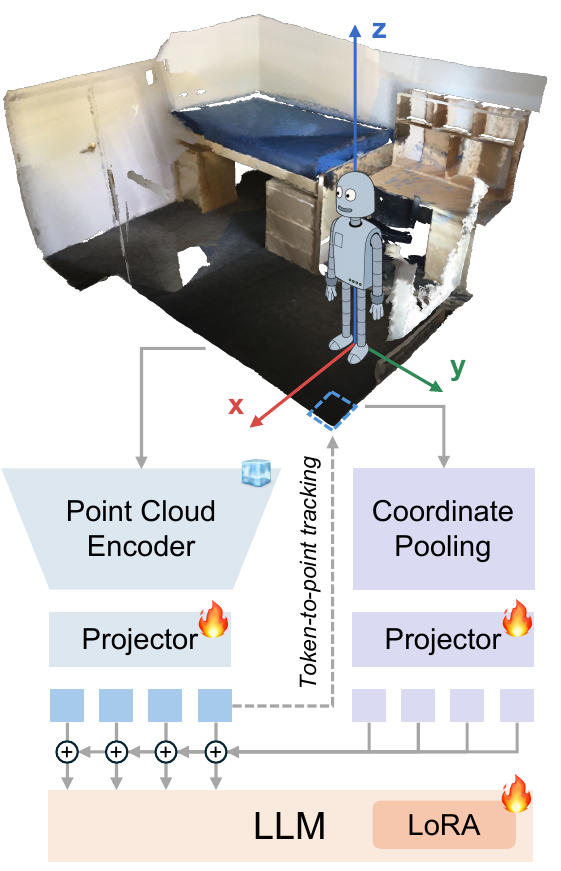}
  \caption{EgoAlign architecture with explicit coordinate fusion.
    The color branch is omitted for clarity.}
  \label{fig:model_architecture}
  \vspace{-15pt}
\end{wrapfigure}

EgoAlign connects a pretrained point encoder to an LLM.
The encoder compresses the transformed cloud into $M\ll N$ semantic tokens $\{\mathbf{z}_j\}_{j=1}^{M}$. 
Although the input cloud is expressed in the agent-centric frame, the compressed semantic tokens do not explicitly preserve the metric location of their supporting points. This information can be attenuated by point-cloud encoders designed to learn robust, translation and rotation invariant geometric representations \citep{wu2025sonata, zhang2026concerto}, yet it is essential for reasoning about the scene relative to the agent.

To address this problem, we infuse metric coordinates explicitly. 
For token $j$, we track the encoder's pooling assignments to obtain $\mathcal{I}_j$, the original input points contributing to that token.
We mean-pool the coordinates and colors of the points in $\mathcal{I}_j$, then project them alongside the semantic feature into the LLM's embedding space:
\begin{equation}
  \mathbf{e}_j = \frac{1}{\sqrt{3}}\left[
    f_{\mathrm{sem}}(\mathbf{z}_j)
    + f_{\mathrm{coord}}\!\left(\operatorname*{avg}_{i\in\mathcal{I}_j}\widetilde{\mathbf{p}}_i\right)
    + f_{\mathrm{rgb}}\!\left(\operatorname*{avg}_{i\in\mathcal{I}_j}\mathbf{c}_i\right)
  \right].
  \label{eq:point_token_fusion}
\end{equation}
We scale the sum by \(1/\sqrt{3}\) to approximately preserve the feature variance when combining three independently projected branches.
The fused point cloud embeddings are then processed by the LLM's attention layers together with the regular text embeddings.

\begin{figure}[t]
  \centering
  \includegraphics[width=\linewidth]{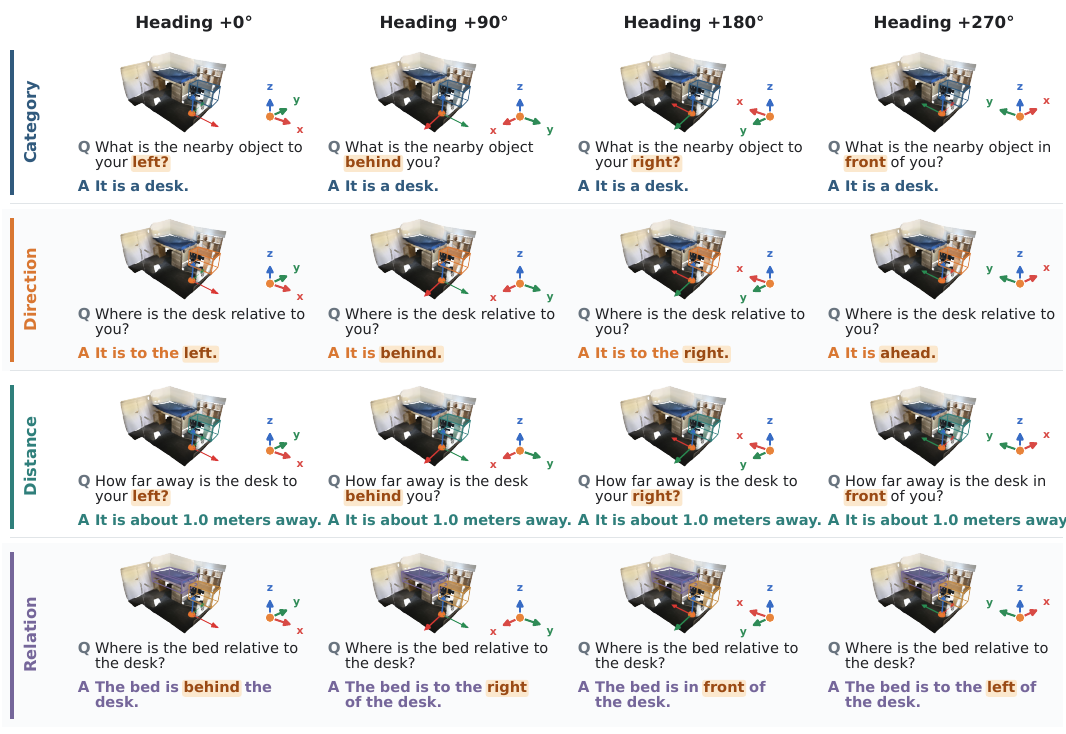}
  \caption{Egocentric alignment supervision at a fixed scene and position.
    Rotating the numerical heading changes agent-relative questions and answers; the scene geometry is unchanged.}
  \label{fig:alignment_examples}
  \vspace{-5pt}
\end{figure}

\subsection{Egocentric alignment}

A suitable representation alone does not ensure that the model learns to use its numerical reference frame. Downstream situated QA also provides insufficient supervision for learning a pose-dependent spatial representation. Besides the linguistic shortcuts present in existing benchmarks~\citep{huang2025unveiling, ma20263d}, the training data contain very limited variation in heading at a fixed agent position. In SQA3D, the 26,623 training questions correspond to 5,308 unique agent positions, of which 5,212 appear with only a single heading. As a result, the model rarely observes how the same scene configuration should be interpreted differently when only the agent orientation changes. We therefore introduce a dedicated egocentric alignment stage that directly supervises the mapping between scene geometry and the agent-centric reference frame. We construct the alignment data from scenes and agent positions in the SQA3D training split, using four complementary tasks:
\begin{itemize}
  \item \textbf{Category} asks the model to identify an object from its color and egocentric location, grounding semantic point tokens to a specified spatial region.
  \item \textbf{Direction} predicts one of eight horizontal directions from the agent to an object, explicitly supervising how object location changes with the agent heading.
  \item \textbf{Distance} estimates the nearest horizontal distance from the agent to an object's box, rounded to 0.1 meters, connecting egocentric coordinates to quantitative spatial language.
  \item \textbf{Relation} describes one object's location relative to another using eight agent-centric horizontal relations together with the viewpoint-invariant \emph{above} and \emph{below} relations.
\end{itemize}

Crucially, each query is expanded into a four-heading counterfactual group. We keep the scene, agent position, and referred objects fixed while instantiating headings at \(0^\circ\), \(90^\circ\), \(180^\circ\), and \(270^\circ\). Thus, the underlying scene content remains unchanged, while the egocentric coordinates and corresponding spatial labels must vary consistently with the numerical heading. This construction provides the heading variation largely missing from downstream QA and directly teaches the model how spatial interpretation should transform with the agent frame. We sample 75K examples for each task, yielding a balanced 300K-example alignment corpus, and jointly train all four tasks with the same autoregressive objective used for downstream QA.
Figure~\ref{fig:alignment_examples} illustrates one four-heading group.

\section{Experiments}
\label{sec:experiments}

\subsection{Experimental setup}

\paragraph{Model.}
EgoAlign couples a Concerto point encoder~\citep{zhang2026concerto} with a Qwen2.5-Instruct language model \citep{qwen2.5}.
We instantiate EgoAlign with the 1.5B, 3B, and 7B variants of Qwen2.5-Instruct. During both egocentric alignment and downstream fine-tuning, the point encoder and base LLM weights remain frozen; we train only the projection layers and LoRA adapters. Optimization and implementation details are provided in Appendix~\ref{app:training}.

\paragraph{Datasets and metrics.}
We evaluate on SQA3D, Real-3DQA, and MSQA. SQA3D contains 33.4K situated questions across 650 ScanNet scenes. Real-3DQA is built on SQA3D to reduce linguistic shortcuts: it retains 1,485 test questions after filtering those answerable without 3D evidence, and augments 750 questions with four agent headings. MSQA contains 251K situated QA pairs across 1,734 scenes from ScanNet, 3RScan, and ARKitScenes. We report exact match (EM) and refined exact match (EM-R) on SQA3D and Real-3DQA. For the four-heading subset of Real-3DQA, we additionally report viewpoint robustness score (VRS), which averages QA accuracy across the four heading variants. On MSQA, we report scores using its official LLM-based evaluator.

\paragraph{Baselines.}
We compare EgoAlign with a broad set of 3D language models, including 3D-LLM~\citep{hong20233d}, LEO~\citep{huang2023embodied}, Chat-Scene~\citep{huang2024chat}, Spartun3D-LLM~\citep{zhang2024spartun3d}, 3D-LLaVA~\citep{deng20253d}, PoseAlign~\citep{liu2026direction}, MSR3D~\citep{linghu2024multi}, SceneCOT~\citep{linghu2026scenecot}, Scene-LLM~\citep{fu2024scene}, LLaVA-3D~\citep{zhu2024llava}, Video-3D LLM~\citep{zheng2025video}, GPT4Scene~\citep{qi2026gpt4scene}, and Qwen-3D~\citep{lin2026qwen}. We distinguish methods that consume 3D scene representations from those solely based on images or videos in the result tables.

\begin{table}[t]
  \centering
  \caption{Results on SQA3D and Real-3DQA. EgoAlign uses Qwen2.5-7B.
    \posekey{} cells indicate methods that incorporate numerical poses. $+$sit./$-$sit. denote with/without situation descriptions.}
  \label{tab:sqa3d_real3dqa_main}
  \small
  \setlength{\tabcolsep}{2.7pt}
  \renewcommand{\arraystretch}{1.09}
  \begin{tabularx}{\linewidth}{@{}lcc*{2}{>{\centering\arraybackslash}X}
    !{\hspace{2.5pt}{\color{tabrule}\vrule width 0.4pt}\hspace{2.5pt}}
    *{2}{>{\centering\arraybackslash}X}
    !{\hspace{2.5pt}{\color{tabrule}\vrule width 0.4pt}\hspace{2.5pt}}
    >{\centering\arraybackslash}X@{}}
    \toprule
    & \multicolumn{2}{c}{Encoder} & \multicolumn{2}{c}{SQA3D} &
      \multicolumn{3}{c}{Real-3DQA} \\
    \cmidrule(lr){2-3}\cmidrule(lr){4-5}\cmidrule(l){6-8}
    Method & Point & Vision & EM & EM-R &
      EM & EM-R & VRS \\
    \midrule
    3D-LLM\pub{NeurIPS 2023} & \cmark & \cmark
      & 47.8 & 49.6 & 7.5 & 10.4 & 9.6 \\
    \cellcolor{poseblue}LEO\pub{ICML 2024} & \cmark & \cmark
      & 49.7 & 52.3 & 14.3 & 19.1 & 14.3 \\
    Chat-Scene\pub{NeurIPS 2024} & \cmark & \cmark
      & 54.4 & 57.2 & 17.0 & 22.1 & 12.9 \\
    \cellcolor{poseblue}Spartun3D-LLM\pub{ICLR 2025} & \cmark & \cmark
      & 54.8 & -- & -- & -- & -- \\
    3D-LLaVA\pub{CVPR 2025} & \cmark & \xmark
      & 54.5 & 56.6 & 19.7 & 24.9 & 14.8 \\
    \cellcolor{poseblue}PoseAlign\pub{CVPR 2026} & \cmark & \xmark
      & 54.2 & 56.9 & 18.3 & 23.2 & 14.9 \\
    \midrule
    Scene-LLM\pub{WACV 2025} & \xmark & \cmark
      & 54.2 & -- & -- & -- & -- \\
    LLaVA-3D\pub{ICCV 2025} & \xmark & \cmark
      & 56.3 & 59.0 & 27.7 & 32.6 & 18.3 \\
    Video-3D LLM\pub{CVPR 2025} & \xmark & \cmark
      & 58.5 & 61.2 & 32.3 & 37.3 & 15.2 \\
    GPT4Scene\pub{ICLR 2026} & \xmark & \cmark
      & \underline{60.6} & \underline{63.3} & \underline{33.1} & 36.9
      & 18.2 \\
    Qwen-3D\pub{ECCV 2026} & \xmark & \cmark
      & 59.6 & -- & -- & -- & -- \\
    \midrule
    \cellcolor{poseblue}\textbf{EgoAlign} ($+$sit.)
      & \cmark & \xmark
      & \textbf{61.2} & \textbf{64.3} & \underline{33.3} & \underline{38.5}
      & \underline{18.6} \\
    \cellcolor{poseblue}\textbf{EgoAlign} ($-$sit.)
      & \cmark & \xmark
      & 60.4 & 63.2 & \textbf{33.7} & \textbf{38.6} & \textbf{19.1} \\
    \bottomrule
  \end{tabularx}
\end{table}

\vspace{-5pt}

\begin{table}[t]
  \centering
  \caption{Results on MSQA. EgoAlign uses Qwen2.5-7B.
    \posekey{} cells indicate methods that explicitly incorporate numerical poses. $+$sit./$-$sit. denote with/without situation descriptions.}
  \label{tab:msqa_main}
  \small
  \setlength{\tabcolsep}{2.8pt}
  \renewcommand{\arraystretch}{1.07}
  \begin{tabularx}{\linewidth}{@{}lcc*{6}{>{\centering\arraybackslash}X}
    !{\hspace{3pt}{\color{tabrule}\vrule width 0.4pt}\hspace{3pt}}
    >{\centering\arraybackslash}X@{}}
    \toprule
    & \multicolumn{2}{c}{Encoder} & \multicolumn{7}{c}{MSQA} \\
    \cmidrule(lr){2-3}\cmidrule(l){4-10}
    Method & Point & Vision & Count. & Exist. & Attr. & Spatial & Nav. & Other & Overall \\
    \midrule
    GPT-4o & -- & -- & 31.2 & 71.4 & \textbf{75.2} & 31.5
      & 36.7 & \textbf{88.0} & 49.7 \\
    \midrule
    \cellcolor{poseblue}LEO\pub{ICML 2024} & \cmark & \cmark & 32.5 & 88.5
      & \underline{58.7} & 44.2 & 39.6 & 81.4 & 54.8 \\
    Chat-Scene\pub{NeurIPS 2024} & \cmark & \cmark & \underline{37.4}
      & \underline{92.0} & 49.0 & 47.0 & 58.3 & 83.7
      & 56.6 \\
    \cellcolor{poseblue}MSR3D\pub{NeurIPS 2024} & \cmark & \cmark
      & 33.5 & 87.5 & 53.7 & 48.9
      & \textbf{61.9} & 75.0 & 56.5 \\
    SceneCOT\pub{ICLR 2026} & \cmark & \cmark & \textbf{47.9} & 82.1 & 49.6
      & 47.2 & 51.6 & 80.3 & 55.6 \\
    \midrule
    \cellcolor{poseblue}\textbf{EgoAlign} ($+$sit.)
      & \cmark & \xmark
      & 36.7 & 91.7 & 51.7 & \underline{51.9} & 59.5
      & \underline{84.1} & \underline{58.5} \\
    \cellcolor{poseblue}\textbf{EgoAlign} ($-$sit.)
      & \cmark & \xmark
      & 37.0 & \textbf{93.0} & 54.7 & \textbf{54.0} & \underline{61.2}
      & 83.1 & \textbf{60.0} \\
    \bottomrule
  \end{tabularx}
\end{table}

\vspace{-5pt}

\subsection{Downstream 3D question answering}

\paragraph{SQA3D and Real-3DQA.}
Table~\ref{tab:sqa3d_real3dqa_main} reports the main results on SQA3D and Real-3DQA using EgoAlign with Qwen2.5-7B. EgoAlign achieves the best overall performance on both benchmarks, reaching 61.2 EM and 64.3 EM-R on SQA3D and outperforming prior methods across all reported Real-3DQA metrics. Notably, strong performance does not require textual situation descriptions: removing them only slightly reduces SQA3D EM from 61.2 to 60.4, while slightly improving performance on Real-3DQA. These results show that EgoAlign can establish an effective egocentric reference frame directly from numerical pose, with situation text serving as optional auxiliary context rather than a necessary representation of the agent state.

\begin{wrapfigure}{r}{0.45\textwidth}
  \vspace{-\intextsep}
  \centering
  \includegraphics[width=\linewidth]{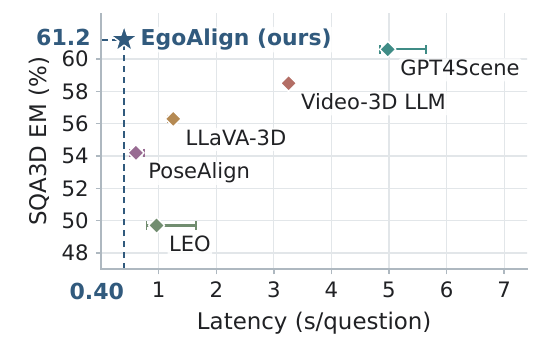}
  \caption{SQA3D exact match versus batch-1 latency.
    Markers and whiskers show median and interquartile. All methods are measured on A40 GPUs.}
  \label{fig:efficiency_sqa3d}
  \vspace{-15pt}
\end{wrapfigure}

\paragraph{Efficiency.}
EgoAlign is also substantially more efficient than existing high-performing models. As shown in Figure~\ref{fig:efficiency_sqa3d}, the 7B model achieves 61.2 SQA3D EM with a batch-1 latency of only 0.40 s/question on an A40 GPU. It is simultaneously faster and more accurate than the compared methods, including both point-based and vision/video-based models. This efficiency comes from operating directly on a compact set of point tokens rather than processing multiple rendered views or video frames.

\paragraph{MSQA.}
Table~\ref{tab:msqa_main} shows similar gains on the larger and more diverse MSQA benchmark. EgoAlign achieves an overall score of 58.5 with situation descriptions and 60.0 without them. The no-situation model performs particularly well on existence and spatial questions, reaching 93.0 and 54.0, respectively. The improvement after removing situation descriptions is consistent with the Real-3DQA results and suggests that direct numerical-pose grounding transfers beyond SQA3D to different datasets and scene sources.

\begin{table}[t]
  \centering
  \caption{Ablations with Qwen2.5-1.5B.
    Models are fine-tuned and evaluated without situation descriptions.
    Distance is measured by MAE.
     $\Delta_{\mathrm{pose}}$ is the decrease in EM under pose shuffling.}
  \label{tab:component_ablation}
  \small
  \setlength{\tabcolsep}{2.8pt}
  \begin{tabularx}{\linewidth}{@{}l*{8}{>{\centering\arraybackslash}X}@{}}
    \toprule
    & \multicolumn{4}{c}{Alignment} & \multicolumn{2}{c}{SQA3D}
    & \multicolumn{2}{c}{Real-3DQA} \\
    \cmidrule(lr){2-5}\cmidrule(lr){6-7}\cmidrule(l){8-9}
    Setting
    & Cat.~$\uparrow$
    & Dir.~$\uparrow$
    & Dist.~$\downarrow$
    & Rel.~$\uparrow$
    & EM~$\uparrow$ & $\Delta_{\mathrm{pose}}$ 
    & EM~$\uparrow$ &  $\Delta_{\mathrm{pose}}$ \\
    \midrule
    \textbf{Full} & \textbf{61.8} & \textbf{59.0} & \textbf{0.47} &
      \textbf{42.2} & \textbf{56.8} & 9.72 & \textbf{30.2} & 10.30 \\
    No alignment & -- & -- & -- & -- & 49.0 & 0.14 & 14.8 & 0.07 \\
    Default coordinates & 28.5 & 12.5 & 1.15 & 11.6 & 48.6 & -- & 16.4 & -- \\
    No coordinate proj. & 49.5 & 38.4 & 0.86 & 34.2 & 55.8 & 6.17 & 25.9 & 6.46 \\
    \midrule
    No category task & 0.3 & 54.0 & 0.73 & 27.2 & 51.3 & 2.13 & 18.1 & 1.41 \\
    No direction task & 47.7 & 23.6 & 1.01 & 30.6 & 53.0 & 3.24 & 23.5 & 5.19 \\
    No distance task & 55.9 & 45.3 & -- & \underline{36.5} & 55.8 & 6.25
      & 23.6 & 7.01 \\
    No relation task & \underline{61.5} & \underline{56.2} & \underline{0.48}
      & 6.8 & \underline{56.0} & 7.72 & \underline{27.8} & 8.16 \\
    Single heading & -- & -- & --
      & -- & 55.6 & 7.13 & 27.0 & 7.28 \\
    \bottomrule
\end{tabularx}
\end{table}

\vspace{-5pt}

\begin{figure}[t]
  \centering
  \includegraphics[width=\linewidth]{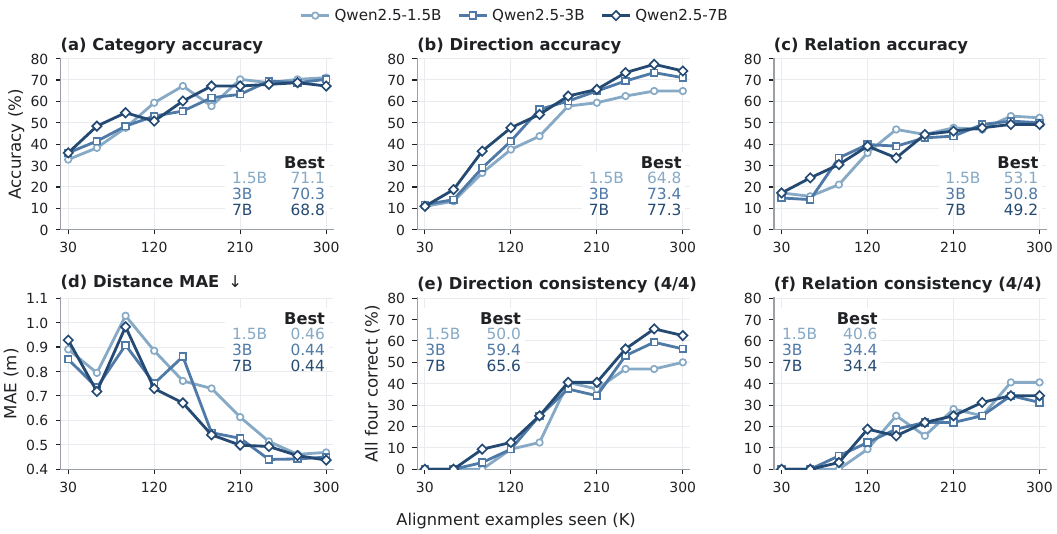}
  \caption{Alignment dynamics across Qwen2.5 model scales.}
  \label{fig:alignment_scaling}
  \vspace{-5pt}
\end{figure}

\subsection{Ablations} Table~\ref{tab:component_ablation} studies the contributions of the agent-centric representation and egocentric alignment using Qwen2.5-1.5B. Removing alignment substantially degrades downstream accuracy and nearly eliminates sensitivity to pose shuffling, showing that downstream QA supervision alone is insufficient to establish a pose-dependent representation. Using the scan's default coordinates similarly harms both alignment and downstream performance. Removing the explicit coordinate projection causes a smaller accuracy drop, but also reduces \(\Delta_{\mathrm{pose}}\), indicating that the agent-centric transform and explicit metric coordinates provide complementary signals for both reasoning accuracy and pose dependence. Alignment dynamics for these variants are provided in Appendix~\ref{app:alignment_ablation_dynamics}.

The four alignment tasks provide complementary supervision. To isolate task composition, each task ablation resamples from the remaining tasks to keep the alignment set fixed at 300K examples. Category supervision is the most critical, while removing direction, distance, or relation causes smaller but consistent losses in downstream accuracy and pose sensitivity. The relatively mild effect of removing relation supervision is expected, as relation prediction is a higher-level task that overlaps with the directional and metric skills learned elsewhere. We also replace the four-heading counterfactual construction with a 300K-example single-heading variant. Although single-heading alignment remains beneficial, it consistently underperforms the full model in both downstream accuracy and \(\Delta_{\mathrm{pose}}\). Thus, single-heading supervision is already useful, while explicit heading variation provides additional supervision for learning a stronger pose-dependent representation.

\section{Analysis}
\label{sec:analysis}

\begin{figure}[t]
  \centering
  \includegraphics[width=\linewidth]{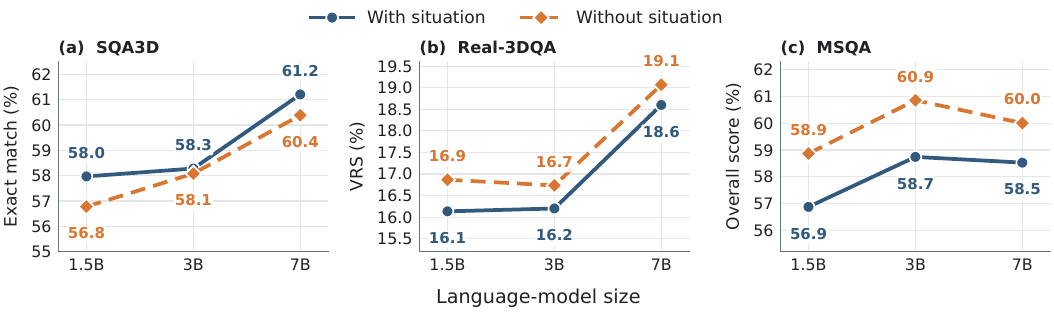}
  \caption{Downstream performance across Qwen2.5 model scales with and without situation descriptions.
    Complete results are in Appendix~\ref{app:model_scale_results}.}
  \label{fig:downstream_scaling}
  \vspace{-5pt}
\end{figure}

\begin{figure}[t]
  \centering
  \includegraphics[width=\linewidth]{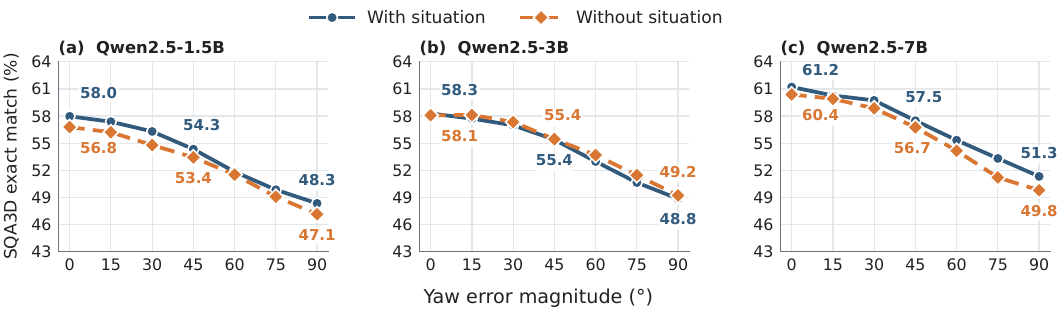}
  \caption{SQA3D exact match under numerical yaw error across Qwen2.5 model scales.
    Perturbation signs are balanced, and $0^\circ$ denotes the clean pose.
    Exact values are in Appendix~\ref{app:yaw_noise_results}.}
  \label{fig:yaw_noise_scaling}
  \vspace{-5pt}
\end{figure}

\subsection{Effect of model scale}
\label{sec:model_scale_analysis}
Figure~\ref{fig:alignment_scaling} examines how egocentric alignment evolves across Qwen2.5 model scales. All three models improve steadily as more alignment examples are observed, although scaling affects individual skills differently. Larger language models benefit direction prediction most clearly, with the best accuracy increasing from 64.8 for 1.5B to 77.3 for 7B, while category and relation accuracy do not consistently improve with model size. Importantly, four-heading consistency also rises markedly during alignment, requiring the model to answer all four counterfactual heading variants correctly. This shows that EgoAlign learns to update its spatial interpretation consistently with the numerical heading. Overall, increasing language-model capacity appears particularly beneficial for heading-dependent directional reasoning, while some grounding skills depend more strongly on the shared 3D representation.

Figure~\ref{fig:downstream_scaling} shows the corresponding downstream results. SQA3D generally improves with model scale, while Real-3DQA also benefits most from the 7B model. MSQA improves from 1.5B to 3B and then largely saturates. Across all three scales, removing situation descriptions improves performance on Real-3DQA and MSQA, whereas SQA3D retains a small benefit from situation text. Thus, EgoAlign does not rely on increasing model scale to compensate for missing situation descriptions; the no-situation setting remains competitive across the full range of model sizes.

\subsection{Sensitivity to pose error}
\label{sec:pose_error_analysis}
A model that genuinely depends on numerical pose should degrade predictably when that pose becomes inaccurate. We therefore perturb the input yaw by increasing angular errors while keeping the scene and question unchanged. Figure~\ref{fig:yaw_noise_scaling} shows a clear degradation across all model scales: for the 7B model, SQA3D EM decreases from 61.2 at the correct pose to 51.3 at a \(90^\circ\) error with situation descriptions, and from 60.4 to 49.8 without them. Similar trends appear for the 1.5B and 3B variants.

The degradation is gradual rather than abrupt, linking model performance directly to the fidelity of the supplied reference frame. Situation descriptions provide only limited robustness to pose corruption: the trends with and without situation text remain closely matched. This indicates that EgoAlign continues to rely primarily on its numerical pose even when linguistic situation context is available.

\subsection{Can alignment supervision transfer to existing architectures?}
\label{sec:alignment_transfer}

\begin{table}[t]
  \centering
  \caption{Transfer of egocentric alignment supervision to existing 7B models.
    All methods use the same 300K alignment examples; SQA3D uses no-situation fine-tuning and evaluation.
    Distance is MAE in meters; $\Delta_{\mathrm{pose}}$ is the decrease in EM under pose shuffling.}
  \label{tab:alignment_transfer}
  \footnotesize
  \setlength{\tabcolsep}{2.0pt}
  \renewcommand{\arraystretch}{1.10}
  \begin{tabularx}{\linewidth}{@{}l*{4}{>{\hsize=.85\hsize\centering\arraybackslash}X}
    !{\hspace{2.5pt}{\color{tabrule}\vrule width 0.4pt}\hspace{2.5pt}}
    *{6}{>{\hsize=1.10\hsize\centering\arraybackslash}X}@{}}
    \toprule
    & \multicolumn{4}{c}{Alignment validation} &
      \multicolumn{3}{c}{SQA3D EM} & \multicolumn{3}{c}{SQA3D EM-R} \\
    \cmidrule(lr){2-5}\cmidrule(lr){6-8}\cmidrule(l){9-11}
    Method & Cat. $\uparrow$ & Dir. $\uparrow$ & Dist. $\downarrow$ &
      Rel. $\uparrow$ & Correct & Shuffled & $\Delta_{\mathrm{pose}}$ &
      Correct & Shuffled & $\Delta_{\mathrm{pose}}$ \\
    \midrule
    LEO & 30.9 & 12.7 & 0.95 & 12.5 &
      46.63 & 46.69 & $-0.06$ & 48.59 & 48.57 & $0.03$ \\
    PoseAlign & 36.3 & 13.0 & 1.09 & 13.2 &
      50.13 & 50.41 & $-0.28$ & 52.46 & 52.71 & $-0.26$ \\
    \midrule
    \textbf{EgoAlign} & \textbf{63.2} & \textbf{61.3} & \textbf{0.42} &
      \textbf{46.6} & \textbf{60.39} & \textbf{49.50} &
      \textbf{$10.88$} & \textbf{63.23} & \textbf{52.12} &
      \textbf{$11.11$} \\
    \bottomrule
  \end{tabularx}
\end{table}

\vspace{-5pt}

One possibility is that our alignment data alone is sufficient to make any pose-conditioned 3D language model pose-sensitive. We test this by applying the same 300K egocentric alignment examples to 7B LEO and PoseAlign variants. As shown in Table~\ref{tab:alignment_transfer}, both models learn the alignment tasks substantially less effectively than EgoAlign. More importantly, their downstream predictions remain insensitive to pose: shuffling the numerical pose changes SQA3D EM by \(+0.06\) for LEO and \(+0.28\) for PoseAlign, compared with a \(-10.88\) point drop for EgoAlign.

These results indicate that alignment supervision alone is insufficient. Effective pose grounding requires a representation that exposes agent-centric metric geometry to the language model together with supervision that teaches how this geometry should be interpreted. EgoAlign is designed around both components, whereas simply applying the same alignment data to existing architectures does not produce comparable pose dependence.

\subsection{Qualitative analysis}
\label{sec:qualitative_analysis}

\begin{figure}[ht]
  \centering
  \includegraphics[width=\linewidth]{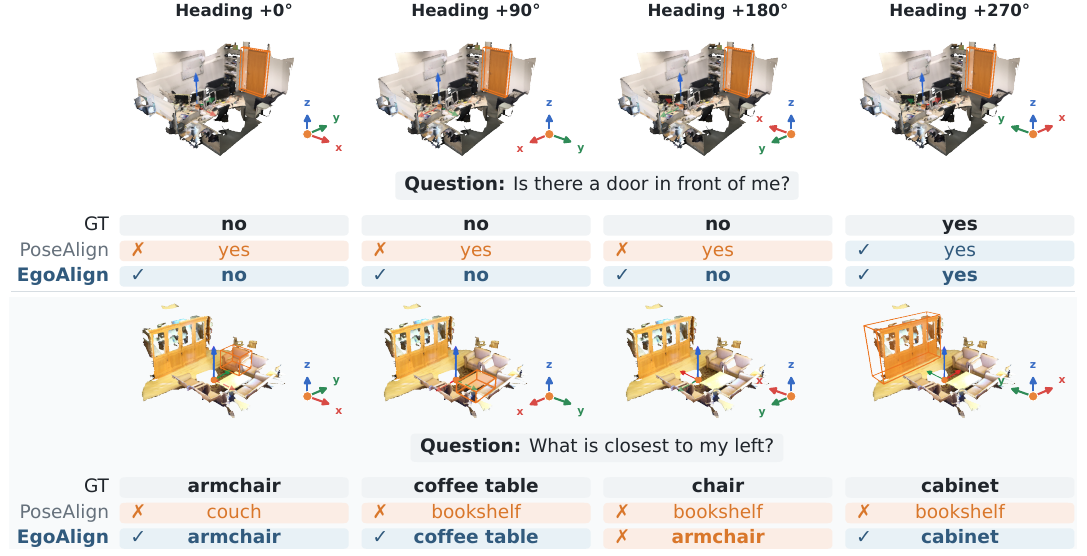}
  \caption{Qualitative Real-3DQA comparisons across four annotated headings.}
  \label{fig:real3dqa_case_study}
  \vspace{-5pt}
\end{figure}

Figure~\ref{fig:real3dqa_case_study} shows two multi-heading examples from Real-3DQA. PoseAlign tends to keep nearly the same answer as the heading changes, while EgoAlign adapts its prediction to the agent frame. In the second example, the dataset answer at \(180^\circ\) is incorrect: the object to the agent's left is an armchair, which is also predicted by EgoAlign. Overall, the examples illustrate that EgoAlign responds consistently to changes in numerical heading rather than producing pose-invariant answers. Additional multi-heading case studies are provided in Appendix~\ref{app:additional_case_studies}.

\section{Conclusion}
\label{sec:conclusion}

We presented EgoAlign, a 3D language model for situated reasoning directly from numerical agent pose. Our experiments reveal pose insensitivity as an important failure mode of existing pose-conditioned models: even when textual situation description is unavailable, their predictions can remain largely unchanged when the numerical pose is corrupted. EgoAlign addresses this issue through an agent-centric point representation that preserves metric coordinates and a dedicated egocentric alignment stage that explicitly teaches how spatial interpretation changes with the agent frame. Across SQA3D, Real-3DQA, and MSQA, EgoAlign achieves excellent performance while remaining efficient, and performs comparably or better without textual situation descriptions. Its predictions also degrade consistently under pose shuffling and increasing yaw error, demonstrating substantially stronger dependence on the supplied reference frame. These results suggest that numerical pose can serve as a direct and effective interface for situated 3D language reasoning, reducing the need to translate an agent’s geometric state into language before reasoning.

\subsection*{AI use statement}
In this work, we used generative AI tools to provide feedback on research framing and experimental design, and to assist in translation and implementing methods. 
We have not used generative AI tools to generate synthetic datasets, formulate or prove mathematical claims, or interpret results, and the remaining required-disclosure tasks are not applicable to this work. 
Additionally, we used generative AI tools for literature search and discovery and polishing the writing for clarity and readability. We reviewed all AI-assisted work, including independently verifying technical claims and citations against the original sources and experimental results. We take responsibility for the final content of this work, including text, claims, or artifacts produced with the aid of generative AI.

\subsection*{Reproducibility statement}
Section~\ref{sec:method} specifies the agent-centric transformation, point-token interface, and egocentric alignment procedure, while Section~\ref{sec:experiments} describes the datasets, evaluation metrics, and experimental setup. The Appendices~\ref{app:pose_shuffle_significance}-- \ref{app:additional-results} provide alignment-data construction details, training hyperparameters, evaluation protocols, statistical tests, and complete numerical results. To facilitate reproduction and future research, we will release our training and evaluation code, processed alignment annotations, model checkpoints, and evaluation outputs upon publication.

\section*{Acknowledgements}
This work was partially supported by the NSF under grants IIS-2533550, IIS-2321504, IIS-2217239, CNS-2426514, and CMMI-2146076, Notre Dame Strategic Framework Research Grant (2025), and Notre Dame Poverty Research Package (2025). Any expressed opinions, findings, and conclusions or recommendations are those of the authors and do not necessarily reflect the views of the sponsors.

\bibliography{references}
\bibliographystyle{iclr2027_conference}

\clearpage
\appendix

\section{Pose-shuffle evaluation and significance}
\label{app:pose_shuffle_significance}

\paragraph{Pose-shuffle evaluation.}
For each SQA3D test scene, we place its distinct annotated agent poses in a deterministic random cycle and replace each pose with its successor.
Questions sharing an original pose receive the same donor pose, and no pose maps to itself.
The donor mapping is shared across methods; the scene, question, answer, and situation description (when present) remain fixed.
We compare correct- and shuffled-pose inputs using the same checkpoint for each method.

Table~\ref{tab:pose_shuffle_significance} reports paired pose-shuffle significance tests with and without situation descriptions. We compute 95\% confidence intervals from per-question paired accuracy differences using a normal approximation, and report two-sided exact McNemar \(p\)-values. For LEO and PoseAlign, pose shuffling causes only negligible changes in EM and EM-R under both settings; all 95\% confidence intervals include zero and none of the differences are statistically significant (\(p\geq .10\)). In contrast, EgoAlign exhibits large and highly significant drops in both settings. With situation descriptions, shuffling the pose reduces EM/EM-R by 9.12/9.58 points, while removing situation descriptions increases the drops to 10.88/11.11 points; all corresponding confidence intervals are far from zero (\(p<10^{-39}\)). These results confirm that EgoAlign is strongly dependent on the supplied numerical pose, whereas the predictions of LEO and PoseAlign remain largely pose-insensitive regardless of whether situation text is available.

\begin{table}[ht]
  \centering
  \caption{Paired SQA3D significance analysis under pose shuffling after fine-tuning with ($+$sit.) or without ($-$sit.) situation descriptions. LEO and PoseAlign are fine-tuned using their official hyper-parameters with officially released checkpoints.}
  \label{tab:pose_shuffle_significance}
  \footnotesize
  \setlength{\tabcolsep}{4.5pt}
  \renewcommand{\arraystretch}{1.06}
  \begin{tabular*}{\linewidth}{@{\extracolsep{\fill}}lllrrrrr@{}}
    \toprule
    Setting & Method & Metric & Correct & Shuffled & Drop & 95\% CI & $p$ \\
    \midrule
    $+$sit. & LEO & EM & 49.73 & 49.82 & $-0.09$ & $[-0.54,0.37]$ & $.80$ \\
             & & EM-R & 52.29 & 52.29 & $0.00$ & $[-0.45,0.45]$ & $1.00$ \\
    \cmidrule(lr){2-8}
             & PoseAlign & EM & 54.22 & 54.39 & $-0.17$ & $[-1.17,0.82]$ & $.78$ \\
             & & EM-R & 56.89 & 56.92 & $-0.03$ & $[-1.05,0.99]$ & $1.00$ \\
    \cmidrule(lr){2-8}
             & \textbf{EgoAlign} & EM & 61.21 & 52.09 & 9.12 & $[7.76,10.49]$ & $<10^{-39}$ \\
             & & EM-R & 64.34 & 54.76 & 9.58 & $[8.21,10.97]$ & $<10^{-41}$ \\
    \midrule
    $-$sit. & LEO & EM & 48.28 & 48.11 & 0.17 & $[-0.43,0.77]$ & $.64$ \\
             & & EM-R & 50.67 & 50.53 & 0.14 & $[-0.48,0.76]$ & $.72$ \\
    \cmidrule(lr){2-8}
             & PoseAlign & EM & 52.00 & 51.55 & 0.45 & $[-0.05,0.96]$ & $.10$ \\
             & & EM-R & 54.50 & 54.08 & 0.43 & $[-0.09,0.94]$ & $.13$ \\
    \cmidrule(lr){2-8}
             & \textbf{EgoAlign} & EM & 60.39 & 49.50 & 10.88 & $[9.35,12.41]$ & $<10^{-42}$ \\
             & & EM-R & 63.23 & 52.12 & 11.11 & $[9.56,12.66]$ & $<10^{-43}$ \\
    \bottomrule
  \end{tabular*}
  \vspace{-5pt}
\end{table}

\section{Training and implementation details}
\label{app:training}

\begin{figure}[ht]
  \centering
  \includegraphics[width=\linewidth]{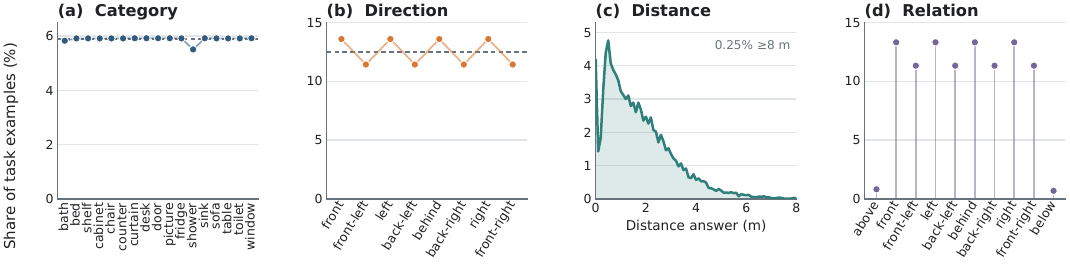}
  \caption{Answer distributions in the alignment set (75K examples per task).
    Dashed lines denote uniform category and direction distributions; distance retains its 0.1~m precision.}
  \label{fig:alignment_data_distribution}
  \vspace{-5pt}
\end{figure}

\paragraph{Alignment data.}
The SQA3D training split provides 518 ScanNet scenes and 4,262 eligible agent positions; scene-disjoint validation uses 65 scenes and 493 positions~\citep{ma2022sqa3d,dai2017scannet}. There is no overlap between scenes used in alignment training and downstream evaluation. 
For each task, we sample 18,750 complete four-heading groups for training and 250 for validation.
ScanNet instance annotations determine the targets but are not model inputs; each example retains the complete scene point cloud.
Category uses 17 object classes after excluding structural surfaces and the catch-all \emph{other furniture} class.
For direction and horizontal relation labels, we take the angle of the target center relative to the agent (or the reference-object center) in the agent-centric $xy$ plane and assign it to the nearest of eight directions at $45^\circ$ intervals. We assign above or below only when the upper object’s center lies within the lower object’s horizontal bounding box and above its top.
Figure~\ref{fig:alignment_data_distribution} reports the resulting answer distributions.

\paragraph{Optimization.}
We initialize the point tower from the official Concerto Large checkpoint and use independent two-layer MLPs for the semantic, coordinate, and color projections.
LoRA adapters have rank 16, $\alpha=32$, and dropout 0.05, and are applied to all attention and MLP projections of the language model.
Alignment runs for one 300K-example epoch with a global batch size of 8, AdamW, a peak learning rate of $5\times10^{-5}$, 3\% linear warmup, cosine decay, BF16, and seed 42.

\paragraph{Point-cloud preprocessing.}
We voxelize each scene at 2.5~cm resolution, retaining one point per occupied voxel.
The point encoder uses voxel indices for sparse operations while retaining the selected points' continuous agent-centric XYZ coordinates.
Before the coordinate projector, token-pooled XYZ is divided elementwise by $(16,16,4)$~m, placing horizontal and vertical offsets on comparable numerical scales while retaining their agent-relative signs.

\paragraph{Task-specific fine-tuning.}
SQA3D uses all 26,623 training questions for three epochs.
MSQA uses 234,639 training examples for one epoch.
Both use a global batch size of 4 and the alignment-stage optimizer, learning rate, warmup ratio, and precision. 
Real-3DQA is directly evaluated using the SQA3D-finetuned checkpoints.

\section{Additional results}
\label{app:additional-results}

\subsection{Alignment dynamics of component ablations}
\label{app:alignment_ablation_dynamics}

\begin{figure}[ht]
  \centering
  \includegraphics[width=\linewidth]{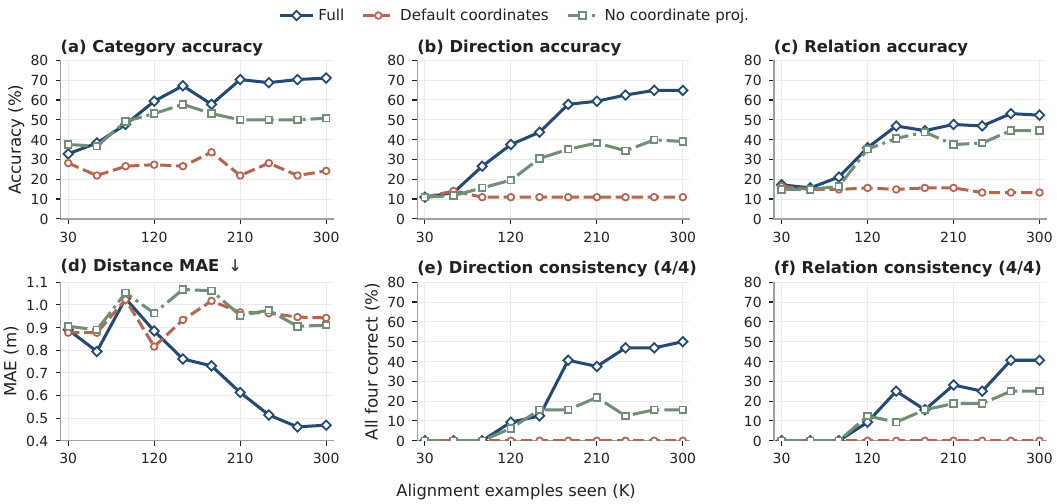}
  \caption{Alignment dynamics for the Qwen2.5-1.5B component ablations.}
  \label{fig:alignment_ablation_dynamics}
  \vspace{-5pt}
\end{figure}

Figure~\ref{fig:alignment_ablation_dynamics} expands the component ablations in Table~\ref{tab:component_ablation} by tracking alignment performance throughout training. In addition to per-example accuracy, we report 4/4 consistency for the direction and relation tasks, defined as the fraction of four-heading counterfactual groups for which all four variants are answered correctly. The full model consistently outperforms both ablations across the four alignment tasks and the consistency metrics. In particular, it reaches substantially higher 4/4 consistency, showing that EgoAlign learns to update its predictions coherently as the numerical heading changes. Using the default scene coordinates largely prevents the model from learning heading-dependent direction and relation predictions, while removing the explicit coordinate projector yields intermediate performance. These results further show that transforming the scene into the agent frame and explicitly exposing metric coordinates to the LLM are complementary.

\subsection{Complete model-scale results}
\label{app:model_scale_results}

\begin{table}[h]
  \centering
  \caption{Full SQA3D and Real-3DQA results across model scales (\%).
    Type columns report VRS within each question type; darker shading indicates higher performance.}
  \label{tab:model_scale_sqa_real}
  \small
  \setlength{\tabcolsep}{2.8pt}
  \renewcommand{\arraystretch}{1.07}
  \begin{tabularx}{0.88\linewidth}{@{}ll*{2}{>{\centering\arraybackslash}X}
    !{\hspace{2.0pt}{\color{tabrule}\vrule width 0.4pt}\hspace{2.0pt}}
    *{2}{>{\centering\arraybackslash}X}
    !{\hspace{1.5pt}{\color{tabrule}\vrule width 0.4pt}\hspace{1.5pt}}
    >{\centering\arraybackslash}X
    !{\hspace{1.5pt}{\color{tabrule}\vrule width 0.4pt}\hspace{1.5pt}}
    *{4}{>{\centering\arraybackslash}X}@{}}
    \toprule
    & & \multicolumn{2}{c}{SQA3D} & \multicolumn{7}{c}{Real-3DQA} \\
    \cmidrule(lr){3-4}\cmidrule(l){5-11}
    Size & Input & EM & EM-R & EM & EM-R & VRS &
      Dist. & Dir. & Count. & Exist. \\
    \midrule
    1.5B & $+$sit. & \scorecell{13}{58.0} & \scorecell{14}{60.6} & \scorecell{18}{30.4} & \scorecell{18}{35.8} & \scorecell{8}{16.1} & \scorecell{8}{12.9} & \scorecell{22}{26.5} & \scorecell{22}{21.7} & \scorecell{27}{61.5} \\
         & $-$sit. & \scorecell{8}{56.8} & \scorecell{8}{59.7} & \scorecell{13}{30.2} & \scorecell{13}{35.3} & \scorecell{22}{16.9} & \scorecell{18}{14.0} & \scorecell{13}{25.4} & \scorecell{13}{19.2} & \scorecell{32}{67.3} \\
    \midrule
    3B   & $+$sit. & \scorecell{22}{58.3} & \scorecell{20}{61.1} & \scorecell{22}{30.9} & \scorecell{22}{36.2} & \scorecell{13}{16.2} & \scorecell{13}{13.1} & \scorecell{32}{27.2} & \scorecell{32}{22.5} & \scorecell{8}{46.2} \\
         & $-$sit. & \scorecell{18}{58.1} & \scorecell{14}{60.6} & \scorecell{8}{28.6} & \scorecell{8}{33.5} & \scorecell{18}{16.7} & \scorecell{22}{14.1} & \scorecell{8}{25.0} & \scorecell{18}{20.8} & \scorecell{13}{53.8} \\
    \midrule
    7B   & $+$sit. & \scorecell{32}{61.2} & \scorecell{32}{64.3} & \scorecell{27}{33.3} & \scorecell{27}{38.5} & \scorecell{27}{18.6} & \scorecell{27}{16.4} & \scorecell{18}{25.9} & \scorecell{8}{18.3} & \scorecell{22}{57.7} \\
         & $-$sit. & \scorecell{27}{60.4} & \scorecell{26}{63.2} & \scorecell{32}{33.7} & \scorecell{32}{38.6} & \scorecell{32}{19.1} & \scorecell{32}{16.5} & \scorecell{32}{27.2} & \scorecell{32}{22.5} & \scorecell{22}{57.7} \\
    \bottomrule
  \end{tabularx}
\end{table}

\begin{table}[ht]
  \centering
  \caption{Full MSQA results across model scales (\%).
    Darker shading indicates higher performance within each column.}
  \label{tab:model_scale_msqa}
  \small
  \setlength{\tabcolsep}{2.8pt}
  \renewcommand{\arraystretch}{1.07}
  \begin{tabularx}{0.88\linewidth}{@{}>{\raggedright\arraybackslash}p{0.09\linewidth}
    >{\raggedright\arraybackslash}p{0.11\linewidth}
    *{6}{>{\centering\arraybackslash}X}
    !{\hspace{3pt}{\color{tabrule}\vrule width 0.4pt}\hspace{3pt}}
    >{\centering\arraybackslash}X@{}}
    \toprule
    & & \multicolumn{7}{c}{MSQA} \\
    \cmidrule(l){3-9}
    Size & Input & Count. & Exist. & Attr. & Spatial & Nav. & Other & Overall \\
    \midrule
    1.5B & $+$sit. & \scorecell{22}{38.2} & \scorecell{13}{91.1} & \scorecell{8}{48.7} & \scorecell{8}{49.0} & \scorecell{8}{57.0} & \scorecell{14}{82.6} & \scorecell{8}{56.9} \\
         & $-$sit. & \scorecell{18}{37.6} & \scorecell{8}{89.8} & \scorecell{13}{51.3} & \scorecell{22}{53.4} & \scorecell{22}{60.1} & \scorecell{32}{84.1} & \scorecell{22}{58.9} \\
    \midrule
    3B   & $+$sit. & \scorecell{27}{39.6} & \scorecell{32}{93.9} & \scorecell{22}{52.0} & \scorecell{13}{51.2} & \scorecell{13}{57.9} & \scorecell{8}{82.0} & \scorecell{18}{58.7} \\
         & $-$sit. & \scorecell{32}{40.5} & \scorecell{18}{91.6} & \scorecell{27}{54.5} & \scorecell{32}{54.6} & \scorecell{32}{62.7} & \scorecell{26}{83.7} & \scorecell{32}{60.9} \\
    \midrule
    7B   & $+$sit. & \scorecell{8}{36.7} & \scorecell{22}{91.7} & \scorecell{18}{51.7} & \scorecell{18}{51.9} & \scorecell{18}{59.5} & \scorecell{32}{84.1} & \scorecell{13}{58.5} \\
         & $-$sit. & \scorecell{13}{37.0} & \scorecell{27}{93.0} & \scorecell{32}{54.7} & \scorecell{27}{54.0} & \scorecell{27}{61.2} & \scorecell{20}{83.1} & \scorecell{27}{60.0} \\
    \bottomrule
  \end{tabularx}
\end{table}

Tables~\ref{tab:model_scale_sqa_real} and~\ref{tab:model_scale_msqa} provide the complete SQA3D, Real-3DQA, and MSQA results underlying the model-scale analysis in the main paper. Scaling generally improves SQA3D and Real-3DQA, with the 7B model obtaining the strongest overall results, whereas MSQA peaks at 3B. Across model sizes, removing situation descriptions remains competitive on SQA3D and consistently improves the overall MSQA score. On Real-3DQA, the no-situation variants also achieve comparable or higher viewpoint-robustness scores, indicating that numerical pose remains effective across model scales.

\subsection{Complete yaw-noise results}
\label{app:yaw_noise_results}
Table~\ref{tab:yaw_noise_complete} reports the exact values corresponding to Figure 7. Performance generally decreases as the numerical heading becomes less accurate across all model scales and both input settings. A \(90^\circ\) yaw error reduces SQA3D EM by 8.9--10.6 points relative to the clean pose, further demonstrating a consistent dependence on the supplied numerical reference frame.

\subsection{Additional qualitative examples}
\label{app:additional_case_studies}
\begin{figure}[ht]
  \centering
  \includegraphics[width=\linewidth]{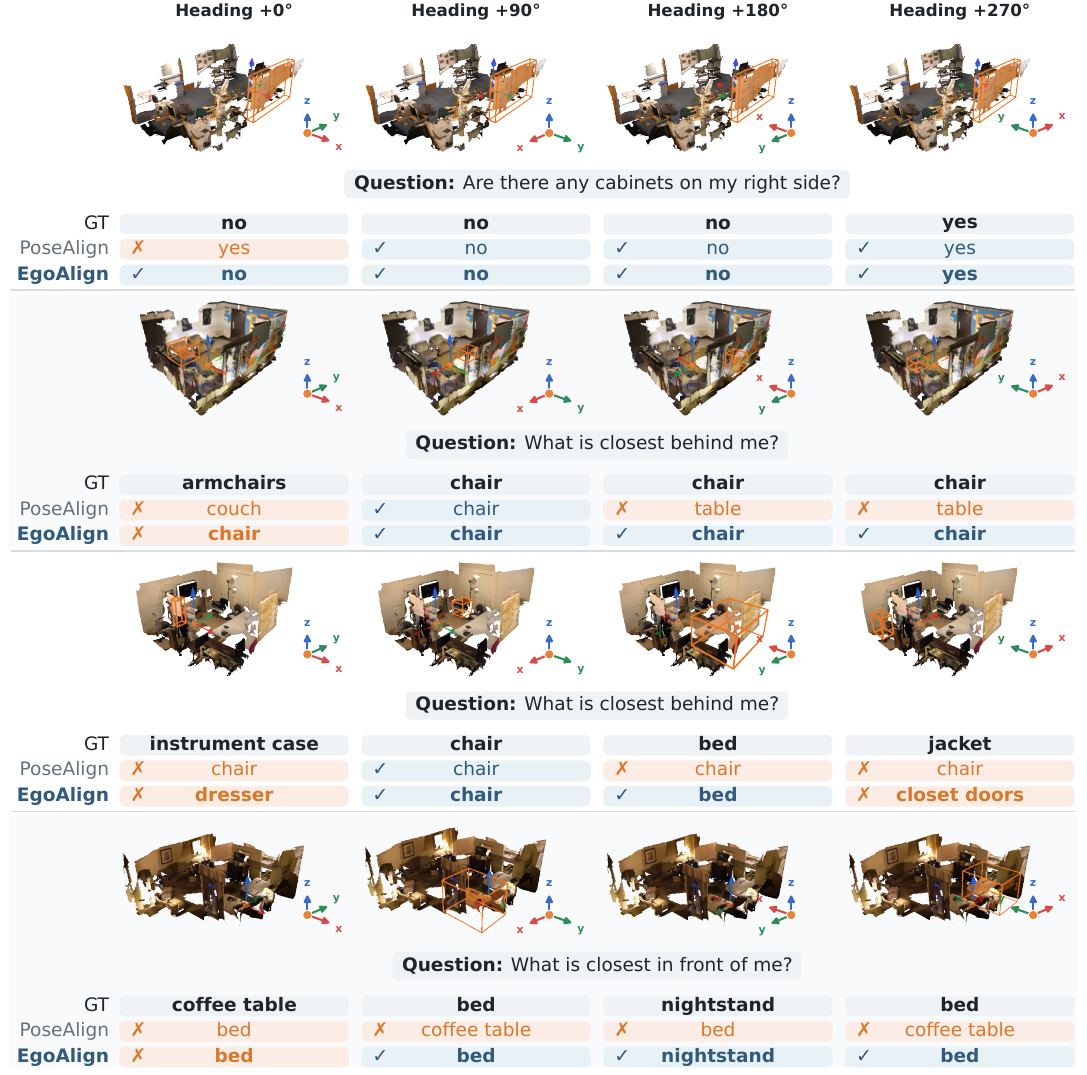}
  \caption{Additional four-heading Real-3DQA examples.
    Within each row, only the agent heading changes.
    Boxes mark matching ScanNet instances in the queried direction.}
  \label{fig:real3dqa_case_study_appendix}
\end{figure}

Figure~\ref{fig:real3dqa_case_study_appendix} provides additional four-heading comparisons on Real-3DQA. Within each row, the scene, agent position, and question remain fixed while only the heading changes. EgoAlign generally adapts its predictions to the changing reference frame, whereas PoseAlign more frequently repeats answers across incompatible headings. EgoAlign is not perfect and still makes errors on some fine-grained closest-object cases, but overall the examples are consistent with the quantitative pose-sensitivity results in the main paper.

\begin{table}[ht]
  \centering
  \caption{SQA3D EM (\%) under numerical yaw perturbations.
    Zero denotes the unperturbed pose; column headings give absolute yaw error in degrees.}
  \label{tab:yaw_noise_complete}
  \small
  \setlength{\tabcolsep}{5.0pt}
  \renewcommand{\arraystretch}{1.06}
  \begin{tabularx}{0.88\linewidth}{@{}>{\raggedright\arraybackslash}p{0.09\linewidth}
    >{\raggedright\arraybackslash}p{0.11\linewidth}
    *{7}{>{\centering\arraybackslash}X}@{}}
    \toprule
    Size & Input & $0^\circ$ & $15^\circ$ & $30^\circ$ & $45^\circ$ &
      $60^\circ$ & $75^\circ$ & $90^\circ$ \\
    \midrule
    1.5B & $+$sit. & 58.0 & 57.4 & 56.3 & 54.3 & 51.8 & 49.8 & 48.3 \\
         & $-$sit. & 56.8 & 56.2 & 54.8 & 53.4 & 51.5 & 49.1 & 47.1 \\
    \midrule
    3B   & $+$sit. & 58.3 & 57.7 & 57.0 & 55.4 & 53.0 & 50.6 & 48.8 \\
         & $-$sit. & 58.1 & 58.1 & 57.3 & 55.4 & 53.7 & 51.5 & 49.2 \\
    \midrule
    7B   & $+$sit. & 61.2 & 60.3 & 59.7 & 57.5 & 55.3 & 53.3 & 51.3 \\
         & $-$sit. & 60.4 & 59.9 & 58.9 & 56.7 & 54.2 & 51.2 & 49.8 \\
    \bottomrule
\end{tabularx}
\end{table}

\end{document}